\documentclass[10pt,twocolumn,letterpaper]{article}

\usepackage[final]{cvpr}      % To produce the REVIEW version
\usepackage{graphicx}
\usepackage{amsmath}
\usepackage{amssymb}
\usepackage{booktabs}
\usepackage[pagebackref,breaklinks,colorlinks]{hyperref}

\usepackage[capitalize]{cleveref}
\crefname{section}{Sec.}{Secs.}
\Crefname{section}{Section}{Sections}
\Crefname{table}{Table}{Tables}
\crefname{table}{Tab.}{Tabs.}

\def\cvprPaperID{659} % *** Enter the CVPR Paper ID here
\def\confName{CVPR}
\def\confYear{2022}

\makeatletter
\newcommand{\printfnsymbol}[1]{%
	\textsuperscript{\@fnsymbol{#1}}%
}
\makeatother

\begin{document}

\title{GenReg: Deep Generative Method for Fast Point Cloud Registration}

	\author{Xiaoshui Huang$^{[1]\thanks{Equal contribution}}$, Zongyi Xu$^{[2]\printfnsymbol{1}}$, Guofeng Mei$^{[3]}$, Sheng Li$^{[4]}$, Jian Zhang$^{[3]}$, Yifan Zuo$^{[4]\thanks{Corresponding author}}$, Yucheng Wang$^{[5]}$ \\
	$[1]$ Image X Institute, University of Sydney,\\
	$[2]$ Chongqing University of Posts and Telecommunications,\\
	$[3]$ GBDTC, FEIT, University of Techonology Sydney, \\
	$[4]$ Jiangxi University of Finance and Economics,
	$[5]$ Xpeng
	% omit the following lines up until the closing ``}
}

\maketitle

%%%%%%%%% ABSTRACT
\begin{abstract}
	Accurate and efficient point cloud registration is a challenge because the noise and a large number of points impact the correspondence search. This challenge is still a remaining research problem since most of the existing methods rely on correspondence search. To solve this challenge, we propose a new data-driven registration algorithm by investigating deep generative neural networks to point cloud registration. Given two point clouds, the motivation is to generate the aligned point clouds directly, which is very useful in many applications like 3D matching and search. We design an end-to-end generative neural network for aligned point clouds generation to achieve this motivation, containing three novel components. Firstly, a point multi-perception layer (MLP) mixer (PointMixer) network is proposed to efficiently maintain both the global and local structure information at multiple levels from the self point clouds. Secondly, a feature interaction module is proposed to fuse information from cross point clouds. Thirdly,  a parallel and differential sample consensus method is proposed to calculate the transformation matrix of the input point clouds based on the generated registration results. The proposed generative neural network is trained in a GAN framework by maintaining the data distribution and structure similarity.  The experiments on both ModelNet40 and 7Scene datasets demonstrate that the proposed algorithm achieves state-of-the-art accuracy and efficiency. Notably, our method reduces $2\times$ in registration error (CD) and $12\times$ running time compared to the state-of-the-art correspondence-based algorithm.
\end{abstract}

\section{Introduction}
Point cloud registration is the cornerstone of numerous 3D computer vision tasks, such as 3D matching, 3D search, and 3D localization \cite{huang2021comprehensive}. Given two point clouds, most existing registration methods \cite{besl1992method,zhou2016fast,huang2017systematic,yew2020rpm,fu2021robust} use a pipeline consisting of four separate stages: keypoint detection, feature extraction, correspondence search and transformation estimation. In the keypoint detection stage, salient points like corner points or all the points are selected as interest points from each point cloud. In the feature extraction phase, local descriptors are extracted for these interest points using their neighbourhood regions. The keypoint detection and feature extraction stages produce two sets of interest points with descriptors. The point-to-point correspondences are later found by nearest neighbour search or more sophisticated matching algorithms.

\begin{figure}[t]
	\includegraphics[width=\linewidth]{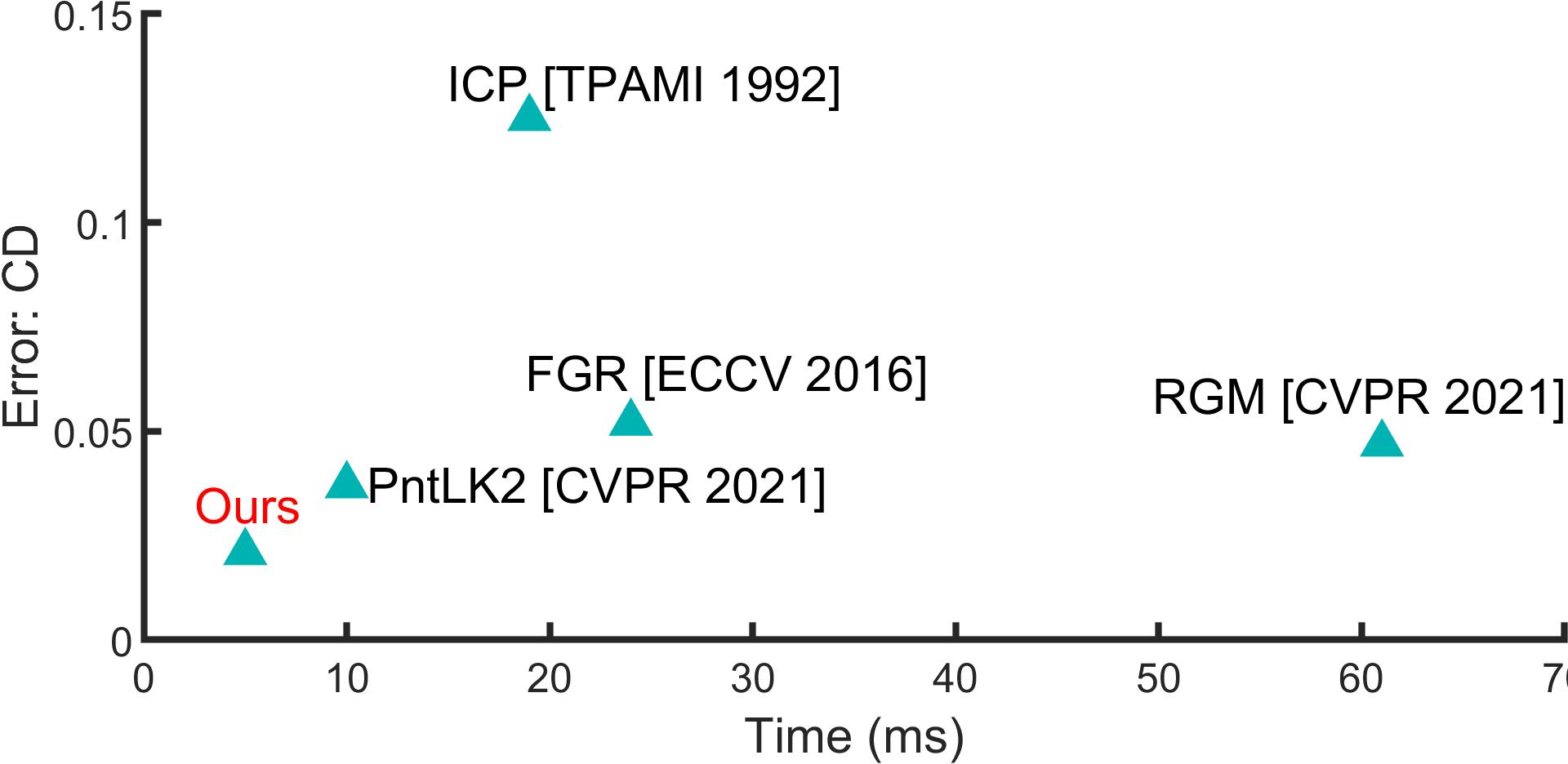}
	\caption{Our GenReg attains lowest registration error (CD: 0.029) and fastest running speed (5 ms) on ModelNet40 test set compared with the state-of-the-art algorithms.}
	\label{f1}
\end{figure}
The use of a keypoint detection reduces matching search space, and the resulting sparse correspondences are sufficient for most tasks, e.g., 3D matching and search. However, a keypoint detector may fail to extract enough interest points that are repeatable between point clouds due to various factors such as repetitive structures, low structures, noise, and outliers \cite{huang2021comprehensive}. Since noise widely exists in point cloud acquisition, repeatable keypoint detection is extremely challenging. Without repeatable interest points, it is a challenge to find correct correspondences even with perfect descriptors. That is the reason why most of the recent methods \cite{zhou2016fast, fu2021robust, huang2021predator} use all the points as interests points. Using all the points will lead to another two challenges. Firstly, finding point-to-point correspondences among all these local descriptors is challenging due to ambiguity structures. Secondly, extracting local descriptors for all the points is time-consuming. Although the recent correspondence-free registration methods \cite{huang2020feature, li2021pointnetlk} align global features without correspondence search, they require an iterative process and omit the local structure information. Both the efficiency and accuracy need to be improved under considerable noise, as shown in our experiments.

We propose a novel data-driven registration algorithm to cast the registration problem into a generation problem. The motivation is that a neural network learns to build the data distribution from two given point clouds and directly generate aligned point clouds by maintaining the same structures with the inputs. Since the learned data distributions represent the statistic models of input point clouds and the algorithm has no iterative process, it is intrinsic to overcome the noise and be fast by directly generating aligned point clouds from the learned data distributions. To achieve this motivation, we need to resolve three critical research questions. Firstly, how to map the input point clouds into an implicit data distribution? Secondly, how to generate aligned point clouds by maintaining the same structures with the inputs? Thirdly, how to estimate the transformation matrix based on the generated registration results?

Four novel components are proposed to resolve the above research questions. Firstly, inspired by MLP-Mixer\cite{tolstikhin2021mlp}, a novel point multi-layer perceptron (MLP) mixer network (PointMixer) is proposed to integrate the token-mixing MLP and channel-mixing MLP. The token-mixing and channel-mixing maintain the global and local structure information respectively in the feature extraction. Compared to other global and local feature networks \cite{qi2017pointnet2,wang2019dynamic}, which use simple concatenation operation, the PointMixer maintain both global and local structure for each point at multiple levels. We train the PointMixer using a generative adversarial network (GAN). Thanks to GAN \cite{goodfellow2020generative}, the extracted features could keep the data distributions of input point clouds. These features contain implicit data distributions and generate aligned point clouds with specific prior knowledge (e.g., input shape). Secondly, a feature interaction module is proposed to fuse information from the cross point clouds further. Thirdly, several loss functions are proposed to train the end-to-end framework to maintain the structural similarity between generated and input point clouds. Fourthly, a parallel and differential sample consensus (PDSAC) is proposed to estimate the transformation matrix of input point clouds in an end-to-end manner, which utilizes the correspondences between generated results and original inputs. 

Specifically,  we express the point cloud registration problem between $P$ and $Q$ in a functional form, as  $P' = \mathcal{F}_\Phi(P, Q)$ and $Q' = \mathcal{F}_\Phi(Q, P)$, where $\mathcal{F}_\Phi$ is our neural network architecture parameterized by $\Phi$, $P'$ and $Q'$ are aligned registration results. $P'$ is the generated point cloud with the prior knowledge of $P$, and $Q'$ is the generated point cloud with the prior knowledge of $Q$.
Our main contributions are:

\begin{itemize}
	\item Based on our best knowledge, this is the first point cloud registration method of directly generating the registration results using a generative neural network. 
	\item A point multi-layer perceptron (MLP) mixer (PointMixer) is proposed to capture point clouds' global and local structure information at multiple levels. 
	\item A feature interaction (FI) module is proposed to fuse the information from cross point clouds.
	\item A parallel and differential sample consensus (PDSAC) is proposed to estimate the transformation matrix in an end-to-end learning framework.
	\item Several loss functions are proposed to train the generative network to maintain structure similarity and implicit data distribution of input point clouds.
	\item We demonstrate that the proposed generative model achieves better accuracy and efficiency.	
\end{itemize}

\section{Related Works}
Based on the registration pipelines, we divide the related works into three categories: correspondence-based, correspondence-free and generative-based  methods.

\subsection{Correspondence-based registration}
The direct idea of point cloud registration is to find the correspondences first and estimate the transformation matrix based on the correspondences \cite{huang2021comprehensive}. The iterative closest point (ICP) \cite{besl1992method} is the widely acknowledged algorithm. Fast global registration (FGR) \cite{zhou2016fast} integrates line constraints to remove the outliers in correspondence search and achieves global optimal in estimating the transformation matrix. Recently, deep learning descriptors have been imported into the field of point cloud registration. FCGF \cite{choy2019fully} proposes a point descriptor based on a UNet architecture and sparse tensor, which has been widely used for correspondence search. Predator \cite{huang2021predator} improves the correspondence search accuracy by importing an attention module after the encoder module of FCGF. RGM \cite{fu2021robust} proposes a learning-based graph matching to improve the accuracy of correspondence search. 

\begin{figure*}
	\includegraphics[width=\linewidth,height=3.5cm]{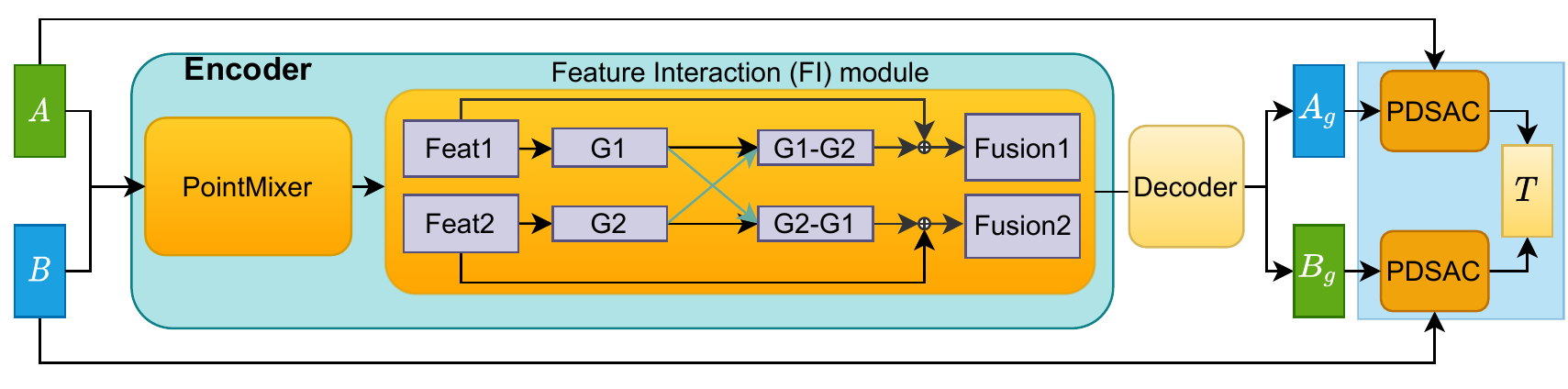}
	\caption{Generative neural network of the proposed GenReg. Our algorithm takes $A$ and $B$ point clouds as input and directly generates the aligned point clouds $A_g$ and $B_g$. The parallel and differential sample consensus (PDSAC) estimates the transformation matrix of input point clouds based on generated results in an end-to-end manner. }
	\label{reggan}
\end{figure*}
However, two limitations existed in this correspondence-based registration pipeline. Firstly, robust correspondence search based on local point descriptors is challenging due to ambiguity/repeated structures, which widely exist in 3D point clouds. Secondly,  this pipeline has high computation cost to calculate point descriptors for all the points.

\subsection{Correspondence-free registration}
To alleviate the above limitations, another registration category is to estimate the transformation matrix without correspondence search. There are two kinds of sub-categories: pose regression and hybrid methods. Pose regression methods extract global features first and directly estimate the pose from the two global features. PCRNet \cite{sarode2019pcrnet} extracts the global features and estimates the transformation matrix from the combined features. OMNet \cite{xu2021omnet} improves this idea by importing an overlapping mask to estimate the overlap regions first. Then, the transformation matrix is regressed by only using the features of overlapped regions. PointNetLK \cite{aoki2019pointnetlk} proposes a hybrid method to combine the deep feature networks and the conventional Lucas-Kanade optimization algorithm to solve the registration problem. FMR \cite{huang2020feature}  improves the PointNetLK by importing an unsupervised branch in the training process. PointNetLK revisit \cite{li2021pointnetlk} proposes a novel Jacobian calculation method to improve the generalization ability.

However, three limitations existed in this correspondence-free registration pipeline. Firstly, the existing registration methods use global features, which weaken the importance of local structures, while the local structures are critical in point cloud registration. Secondly, this pipeline extracts the global feature separately without information interaction between two aligned point clouds. This separated feature learning strategy omit the benefits of combining its aligned point cloud in neural network training. Thirdly, these methods require iterative process to get promising results. The iterative process is time-consuming.

\subsection{Generative-based registration}
To overcome the above limitations in correspondence-free methods, another category applies generative adversarial networks (GAN) \cite{goodfellow2014generative} to solve registration problem. \cite{mahapatra2018deformable} proposes a GAN-based registration method to solve multi-modal medical deformable image registration. However, the algorithm in \cite{mahapatra2018deformable} is designed for medical deformable image registration and faces several challenges by direct transfer to point cloud registration. For example, feature extraction and network training. 

Inspired by \cite{mahapatra2018deformable}, we open up a new research direction in 3D point cloud registration by using generative neural network.
The advantages of the proposed algorithm are two points. Firstly, the proposed method would be much robust to noise that is widely existed in the data acquisition process. Secondly, the overall registration speed is fast since the proposed method runs forward once without an iterative process.

\section{The proposed algorithm: GenReg}
The proposed \textbf{gen}erative-based \textbf{reg}istration (GenReg) algorithm includes three novel modules: 1) a point multi-layer perception mixer (\textit{PointMixer}) module is proposed to keep both global and local structure information at multiple levels for each point; 2) a feature interaction module aims to fuse the global structure information from cross point cloud; 3) a parallel and differential sample consensus (PDSAC) method is proposed to estimate the transformation matrix in an end-to-end manner for the input point clouds. Figure \ref{reggan} shows the overall framework.

\subsection{PointMixer}
\begin{figure}[h]
	\includegraphics[width=\linewidth]{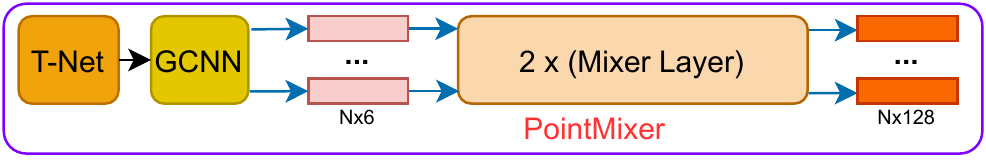}
	\caption{The proposed PointMixer module.}
	\label{pointMixer}
\end{figure}

Inspired by the MLP-Mixer \cite{tolstikhin2021mlp}, we use MLPs operation to mix (1) different channels of a spatial location to keep local information (channel-mixing),  and (2) different spatial points of a channel to keep global information (token-mixing). A mixer layer contains two MLPs to integrate the above two mixings. In our generative neural network, a new \textit{PointMixer}  is proposed to keep both global and local structure information for each point from the self point cloud. Our proposed \textit{PointMixer} includes a T-Net, a graph convolution neural network (GCNN) module and two mixer layers. Compared to other global and local feature networks \cite{qi2017pointnet2,wang2019dynamic}, the PointMixer maintains both global and local structure information at multiple levels. Figure \ref{pointMixer} visually shows the \textit{PointMixer} module.

\textbf{T-Net.}
The goal of T-Net is to normalize the pose difference for registration result generation. The input of T-Net is point cloud (e.g., $A$ or $B$). Inspired by MaskNet\cite{sarode2020masknet}, the T-Net first extracts the global features of two input point clouds and second uses the concatenated global features to learn six freedom parameters (three for rotation angles and three for translation). Then, the six parameters are converted into a $4\times4$ rigid transformation matrix to transform one of the input point clouds. Therefore, the output of T-Net is transformed point cloud (e.g., $A_T$ or $B_T$). $[A_T,1] = [A,1]f_T(A)$ where $f_T$ is a function learned by T-Net. The details of the network are shown in the supplement.

\textbf{GCNN}. The GCNN module extracts initial local features for the input point clouds and prepares input features for the following MLP layers. Following DGCNN \cite{wang2019dynamic}, our GCNN module contains a graph construction layer, a convolution layer and \textit{LeakeyRelu} activation function. 

\textbf{Mixer layers.} The mixer layers will take the graph feature $R^{N\times6}$ as input and output a point feature $R^{N\times 128}$.  The mixer layer extracts the feature for each point by performing token-mixing MLP first and then channel-mixing MLP. The token-mixing MLPs perform on the spatial point dimension to keep global information. By contrast, the channel-mixing MLPs performs on channels of each point to maintain the local information for each point. Each MLP contains two fully connected layers. As our \emph{PointMixer} contains two mixer layers, the final feature maintain global and local structure information at multiple levels.

\subsection{Feature interaction (FI) module} The \emph{FI} module aims to integrate structure information from cross point clouds. 
Specifically, we apply max-pooling to the channel dimension of point features $Feat1$ and $Feat2$ to get the distinct features $G1$ and $G2$. Then, we calculate the difference between these distinct features. Finally, we concatenate the feature difference with point features to get the fusion features $Fusion1$ and $Fusion2$.
Figure \ref{reggan} shows the diagram of our \emph{FI} module.

\subsection{Parallel and differentiable sample consensus}
Although our method can directly generate the aligned point clouds and can be applied to solve 3D matching and retrieve applications, some other applications, like 3D localization and pose estimation, need to know the transformation of two input point clouds.
In the above generative method, the correspondence relationships between generated and real point clouds have been maintained by utilizing our proposed loss functions. Therefore, the transformation matrix can be estimated by utilizing these correspondences. 

In this paper, inspired by \cite{brachmann2017dsac}, we propose a parallel and differential sample consensus (PDSAC) to estimate the transformation so that the whole neural network can be trained in an end-to-end manner. Given a set of one-to-one correspondences $H=(h_1, ..., h_N)$ from $A$ (original input point cloud) and $A_g$ (generated point cloud), we randomly sample $m$ minimal sets at one-time as the hypothesis inlier sets pool $M=(x_1, x_2, ..., x_m)$, in which each element contains $k$ pair correspondences ($e.g.$ $k=4$). Therefore, we could calculate transformation matrix set $T_M=(T_1, T_2, ..., T_m)$ that correspond to each element in $M$  by single value decomposition in a parallel manner. Finally, the projection error of applying the estimated transformation is calculated, and the transformation with minimum projection error $T_A=\min{\sum_{i\in m}} \Arrowvert T_iA- A_g \Arrowvert_2$ is selected as the final predicted transformation matrix. The same for the transformation estimation between $B$ and $B_g$. 

In summary, given estimated correspondences between $A$ and $A_g$, $B$ and $B_g$, there exists following relationships:
\begin{eqnarray}
	\begin{aligned}
		&[A_g,1] = [A,1]T_A\\
		&[B_g,1] = [B,1]T_B
	\end{aligned}
	\label{eqform}
\end{eqnarray}
where $T_A$ and $T_B \in \mathbb{R}^{4\times4}$ could be estimated by utilizing the proposed PDSAC algorithm. $[A,1] \in \mathbb{R}^{N\times4}$ represents the homogeneous coordinates.  Because the $A_g$ and $B_g$ are the generated aligned point clouds, the final transformation matrix of input point clouds $A$ and $B$ can be calculated as below:

\begin{eqnarray}
	T_{est} = (T_A)^{-1}T_B
\end{eqnarray}
where $T_{est}$ is an estimated $4\times4$ rigid transformation between two input point clouds, $[A,1] = [B,1]T_{est}$.

\subsection{Loss functions}
Network training aims to train the generative neural networks to embed the data distribution and generate aligned point clouds in a unified coordinate system. In this section, several loss functions are proposed to achieve this goal.

\subsubsection{Absolute loss} This loss aims to train the generated point cloud should be its aligned point cloud. As shown in Figure \ref{reggan}, the generated point cloud \textit{$A_g$} is aligned with point cloud \textit{B} and generated point cloud \textit{$B_g$} is aligned with point cloud \textit{A}. Their difference is the absolute loss. Since partial overlap widely exists in the registration problem, we use earth mover distance (EMD) to estimate the absolute loss. Mathematically,

\begin{eqnarray}
	l_{abs} =  	EMD(A, B_g)+ 	EMD(B, A_g) 
\end{eqnarray}
where $EMD(X, Y)= \min_{\phi:X \rightarrow Y}\sum_{x\in X} \Arrowvert x - \phi(x) \Arrowvert_2$. The EMD relies on solving an optimization problem that finds a one-to-one bijection mapping $\phi: X \rightarrow Y$, thus only applicable when $\arrowvert Y\arrowvert = \arrowvert X\arrowvert$. The pair-wise distances are then calculated between $x$ and $\phi(x)$.

\subsubsection{Relative loss} 
This loss aims to train the neural network to keep the relative structure similarity between original point cloud $A$ and generated point cloud $A_g$,  original point cloud $B$ and generated point cloud $B_g$. The relative loss is calculated as:
\begin{eqnarray}
	l_{relative} = MAE(E_A, E_{A_g})+ MAE(E_B, E_{B_g})
\end{eqnarray}
where $E_A$, $E_{A_g}$, $E_B$,$E_{B_g}$ are the edge sets, calculating by $E = |P_i- P_{(i+1)\%(N-1)}|_2$, $N$ is the total point number of point cloud $P$. $MAE$ presents the mean absolution error. 

\subsubsection{Cycle consistency loss} 
A network may arbitrarily transform the input point cloud to match the distribution of the target domain. Cycle consistency loss ensures that for each point cloud $A$ the reverse deformation should bring $A$ back to the original point cloud, i.e. $A \rightarrow G(A)=A_g \rightarrow G(F(A)) \simeq A$. The following cycle consistency loss could achieve this goal:

\begin{eqnarray}
	l_{cyc} = EMD(F(G(A)), A) + EMD(F(G(B)), B)
\end{eqnarray}
where $F$ and $G$ are two functions that share the same encoder parameters. 

\subsubsection{Adversarial loss} 
Since the point cloud may contain noise and outliers, the distributions are more reliable information to maintain. Since the generative adversarial network is superior in maintaining distributions, we import adversarial loss in our network training. 

\begin{eqnarray}
	l_{adv} = E_{x\in A}[logD_A(x)] + E_{y\in B}[log(1-D_A(G(y)))]
\end{eqnarray}
where $G$ is the mapping function represent the encoder. $D$ represents the discriminator module.

\subsubsection{Transformation loss} 
Following \cite{huang2020feature,aoki2019pointnetlk}, the transformation loss $l_{T}$  aims to calculate the difference between the estimated transformation of PDSAC and ground truth. $l_T = \|T_{est}T_{gt}^{-1}-I\|_2$. 

Therefore, the total training loss is:
\begin{eqnarray}
	loss = l_{abs} + l_{relative} + l_{cyc} + 0.01*l_{adv}  + l_{T}
\end{eqnarray}

%\subsection{Implementation details} During the training process, following WGAN-GP \cite{gulrajani2017improved}, we add gradient penalty to adversarial loss. Also, the adversarial network is updated every step, and the generative network is updated every five epochs. This training strategy aims to keep the adversarial network more reliable parameters and make the generative network convergent more reliable.
%The discriminator consists of three fully connected layers and activates with the LeakyReLu function. Please look at our supplement for more details.

\section{Experiments}
To evaluate the proposed generative-based point cloud registration method, we evaluate and compare it with the related works on ModelNet40 and 7Scenes.

\subsection{Datasets}
\textbf{ModelNet40}\cite{wu20153d} is a CAD model dataset for evaluating point cloud registration algorithms. Following \cite{huang2020feature,fu2021robust}, the dataset is divided into two parts: 20 categories for training and same-category testing, and another 20 categories for cross-category testing. In the same-category experiments, we split the dataset into $8:2$, where $80\%$ are used for training and $20\%$ for testing. We uniformly sample two times to get two point clouds and normalize them into a unit box at the origin $[0, 1]^3$. This two-time sampling strategy is to simulate the noisy data in the real-world data acquisition process. One of the two sampling point clouds is the source, and the other is the target. Then, the target point cloud is performed a rigid transformation to simulate the pose difference in the real world. During the training and test processes, the rigid transformation $T_{gt}$ is randomly generated with rotation $[0, 45^\circ]$ with arbitrarily chosen axes and translation $[0, 0.8]$.  

\textbf{7Scene \cite{shotton2013scene}} contains seven scenes that are captured by a Kinect RGB-D camera in the indoor environment. This dataset aims to demonstrate the registration performance on real-world data. Following FMR \cite{huang2020feature}, we divide into 296 scans for training and 57 scans for testing. We also uniformly sample two times from the original point clouds and conduct a rigid transformation for one of the two samples. The rotation is initialized in the range of $[0, 45^\circ]$, and the translation is initialized in $[0, 0.8]$ for training and testing.

\subsection{Baselines}

We compare the performance with two kinds of methods, correspondence-based and correspondence-free registration algorithms. {Correspondence-based methods:} include the conventional correspondence-based methods ICP \cite{besl1992method}, fast global registration (FGR) \cite{zhou2016fast}, as well as latest  robust graph matching (RGM) \cite{fu2021robust}.
{Correspondence-free method:} includes the latest PointNetLK revisit (PntLK2) \cite{li2021pointnetlk}.

\begin{figure*}[h]
	\includegraphics[width=\linewidth,height=6.0cm]{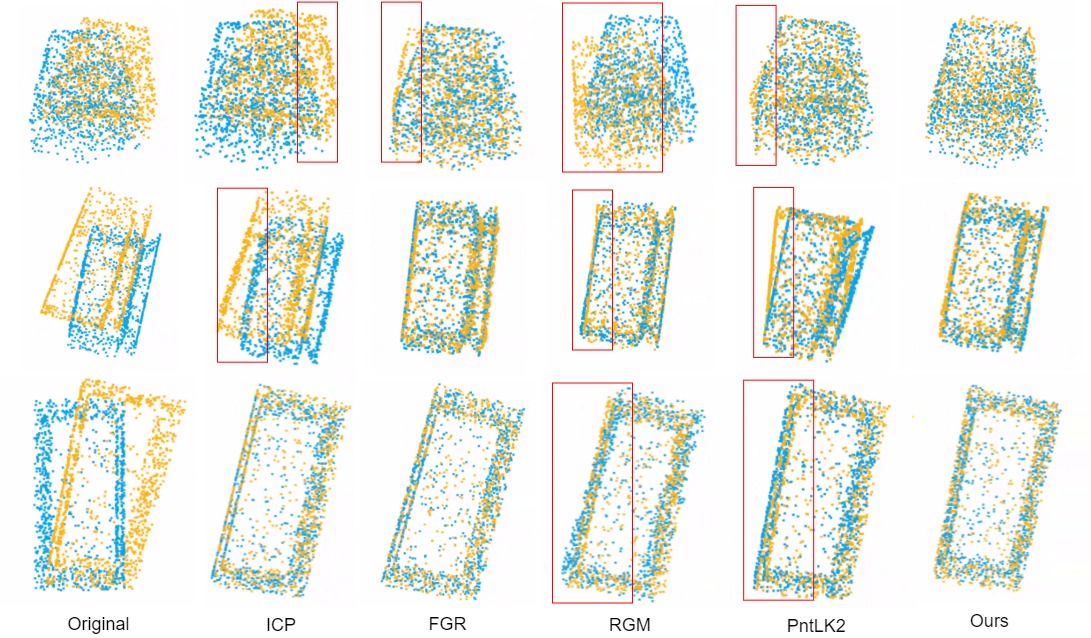}
	\caption{Visual comparison examples. The red boxes show that the proposed method achieves better registration quality.}
	\label{modelnet}
\end{figure*}
\textbf{Evaluation Criteria}. The proposed algorithm directly generates the registration results and estimates the transformation matrix of original point clouds. To evaluate the quality of generated registration results, inspired by \cite{groueix2018papier}, the chamfer distance (CD) between generated point clouds is calculated to evaluate how well the generated point clouds are aligned to each other. The CD is calculated on their aligned point clouds for compared methods. A lower CD means better registration accuracy.  To evaluate the accuracy of transformation matrix estimation, inspired by \cite{xu2021omnet}, the rotation error (RE) and translation error (TE) are calculated as mean absolute errors between estimated rotation angle $\theta_{est}$ and ground truth $\theta_{gt}$, estimated translation  $t_{est}$ and ground truth $t_{gt}$. The unit of rotation angle and time is degree ($^\circ$) and second (s) respectively.

\subsection{Performance on ModelNet40}
\subsubsection{Compare with registration methods}
% 1. original data(two sampling), noise, unseen .
Table \ref{twosamples} shows registration results on the ModelNet40. Our GenReg obtains better registration accuracy and faster registration speed than other compared methods. The lower CD illustrates our better-aligned results than the compared methods. The lower RE and TE demonstrate our better transformation estimation accuracy. Figure \ref{modelnet} visually shows our generated registration results compared with others. The red boxes show that the proposed method obtains better registration accuracy. 

\begin{table}[h]
	\begin{center}
		%\begin{tabular}{p{1.5cm}p{1cm}p{1cm}p{1cm}p{1cm}p{1cm}p{1cm}}
		\begin{tabular}{p{1.8cm}|p{0.8cm}p{0.8cm}p{0.8cm}|p{0.8cm}}\hline			 
			Method       & CD  & RE      & TE  & Time\\
			\hline
			ICP \cite{besl1992method} 	     &0.126 &8.709  &0.116 &0.18\\
			FGR \cite{zhou2016fast} 	     &0.052 &5.565  &0.019 &0.23\\ 
			% DeepGMR      &5.748  &0.012  &0.060 &91.43\% &0.03\\
			RGM \cite{fu2021robust}          &0.044  &2.667  &0.007 &0.61\\ \hline
			%FMR          &0.037 &99.83\% &0.11 &0.745  &0.004\\ 
			PntLK2\cite{li2021pointnetlk}      &0.043 &1.962  &0.009 &0.12\\ \hline
			GenReg       &\bf0.029 &\bf 1.565  &\bf0.006 &\bf0.05\\
			\hline
		\end{tabular}
	\end{center}
	\caption{Performance of two sampling on ModelNet40. }
	\label{twosamples}
\end{table}

Specifically, our method improves two times accuracy in CD and more than ten times speed improvement than the state-of-the-art correspondence-based method RGM. The reason is that the correspondence-based algorithms require estimating correspondences while there are limited high-quality correspondences in the noisy data. Compared to the correspondence-free method, we obtain better registration accuracy and two times faster. The reason is that the state-of-the-art correspondence-free method utilizes the global features and requires running the network inference iteratively ten times. In contrast, the proposed method learns the implicit data distribution using both global and local structure information and generates the registration results directly with one-time network inference. Regarding the transformation estimation of original point clouds, the proposed algorithm demonstrates better accuracy than the compared methods thanks to the proposed PDSAC method.

\begin{table}[h]
	\begin{center}
		%\begin{tabular}{p{1.5cm}p{1cm}p{1cm}p{1cm}p{1cm}p{1cm}p{1cm}}
		\begin{tabular}{p{1.8cm}|p{0.8cm}p{0.8cm}p{0.8cm}|p{0.8cm}}\hline			 
			Method       & CD   & RE    & TE &Time\\ \hline
			ICP \cite{besl1992method} 	     &0.125 &8.733  &0.116 &0.19 \\
			FGR \cite{zhou2016fast}	     &0.052 &5.645  &0.019  &0.24\\ 
			%DeepGMR      &5.054  &0.011  &0.056 &94.01\% &0.03\\
			RGM \cite{fu2021robust}         &0.048   &2.746  &0.009  &0.61\\ \hline
			%FMR          &0.037 &99.83\% &0.10  &0.750  &0.005  \\ 
			PntLK2 \cite{li2021pointnetlk}    &0.043   &1.994  &0.009  &0.11 \\ \hline
			GenReg       &\bf0.029 &\bf1.723  &\bf0.008  &\bf0.05\\ \hline
		\end{tabular}
	\end{center}
	\caption{Performance on point clouds with Gaussian noise.}
	\label{noise}
\end{table}

\textbf{Noise.} In addition, we add Gaussian noise to one of the point clouds to evaluate the robustness. The Gaussian noise is sampled from N (0, 0.01) and independently added to each point in the originally sampled point clouds. Table \ref{noise} shows that the proposed method achieves better registration quality (lower CD), better transformation estimation (lower RE and TE), which demonstrates the proposed method can be high robustness on noise. The registration results have not changed much from the results in Table \ref{twosamples}. The reason is that the two-sampling strategy will generate noise, and this experiment will add more noise to the input point clouds. 

\textbf{Generalization ability.} To demonstrate the generalization ability, the proposed registration method is also evaluated on the cross-category CAD models. Table \ref{unseen} shows that the proposed method can still obtain better accuracy in CD, RE, TE and much faster registration speed. This experiment demonstrates that the proposed method can generalize well in the unseen CAD models.

\begin{table}[h]
	\begin{center}
		%\begin{tabular}{p{1.5cm}p{1cm}p{1cm}p{1cm}p{1cm}p{1cm}p{1cm}}
		\begin{tabular}{p{1.8cm}|p{0.8cm}p{0.8cm}p{0.8cm}|p{0.8cm}}\hline			 
			Method       & CD   & RE    & TE   &Time  \\ \hline
			ICP \cite{besl1992method} 	     &0.125  &8.555  &0.118 &0.18  \\
			FGR \cite{zhou2016fast} 	     &0.055  &4.144  &0.016  &0.28 \\ 
			% DeepGMR      &6.439  &0.013  &0.070 &90.21\% &0.03\\
			RGM \cite{fu2021robust}          &0.052  &2.996  &0.009  &0.59\\ \hline
			%FMR          &0.042 &99.92\% &0.11 &0.612  &0.005  \\ 
			PntLK2 \cite{li2021pointnetlk}      &0.051  &2.157  &0.011 &0.12 \\ \hline
			GenReg       &\bf0.033  &\bf2.093  &\bf0.008 &\bf0.05 \\ \hline
		\end{tabular}
	\end{center}
	\caption{Performance on unseen categories point clouds.}
	\label{unseen}
\end{table}

\begin{figure}[h]
	\includegraphics[width=\linewidth]{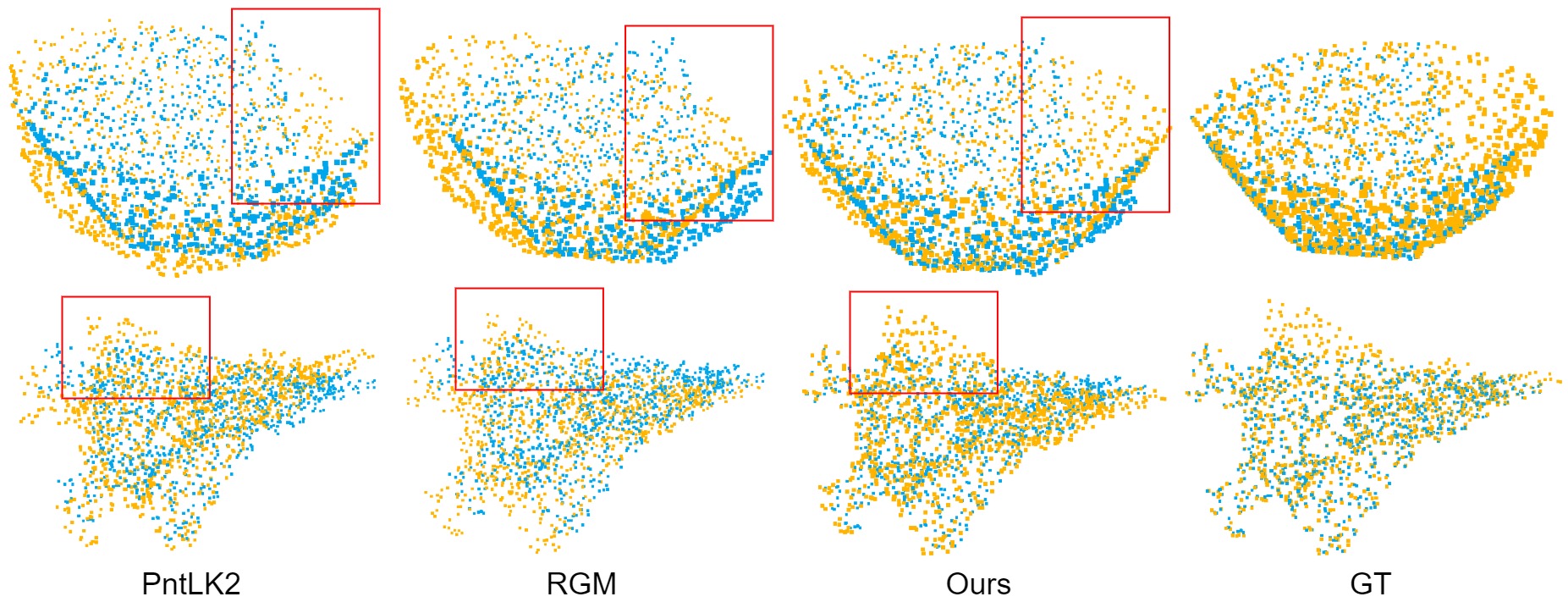}
	\caption{Visual comparison on partial overlapped point clouds. The red box shows our method achieves better accuracy on the partial overlapped parts.}
	\label{partial_fig}
\end{figure}

\textbf{Partial overlap problem.} Partial overlap is a challenging problem in the point cloud registration. To demonstrate the proposed algorithm on the partial overlap problem, we remove the first $20\%$ points based on descending order of X-axis coordinates on one of the sampled point clouds to simulate the partial overlap problem. Then, we run all the methods on these data to compare the performance on partial-to-partial registration. Table \ref{partial} shows that the proposed algorithm achieves better registration quality (lower CD), better transformation estimation accuracy (lower RE and TE) and faster speed. Figure \ref{partial_fig} shows visual comparison examples with the recent state-of-the-art methods. The proposed method achieves visually better registration results on the original point clouds.  This experiment demonstrates that generative-based registration is a new promising way to solve the partial-to-partial registration by learning the implicit data distributions.
\begin{table}[h]
	\begin{center}
		%\begin{tabular}{p{1.5cm}p{1cm}p{1cm}p{1cm}p{1cm}p{1cm}p{1cm}}
		\begin{tabular}{p{1.8cm}|p{0.8cm}p{0.8cm}p{0.8cm}|p{0.8cm}}\hline			 
			Method       & CD   & RE    & TE   &Time  \\ \hline
			ICP \cite{besl1992method} 	     &0.130  &10.256  &0.123 &0.18  \\
			FGR \cite{zhou2016fast} 	     &0.076  &10.507  &0.027  &0.28 \\ 
			% DeepGMR      &6.439  &0.013  &0.070 &90.21\% &0.03\\
			RGM \cite{fu2021robust}          &0.067  &3.522  &0.033  &0.59\\ \hline
			%FMR          &0.042 &99.92\% &0.11 &0.612  &0.005  \\ 
			PntLK2 \cite{li2021pointnetlk}      &0.077  &5.172  &0.031 &0.12 \\ \hline
			GenReg       &\bf0.065 &\bf 3.069  &\bf0.031 &\bf0.05 \\ \hline
		\end{tabular}
	\end{center}
	\caption{Performance on partial overlapped point clouds.}
	\label{partial}
\end{table}

\subsection{Ablation studies}
The critical novelties of the proposed algorithm are the novel \emph{PointMixer} and parallel differential sample consensus (PDSAC) modules. This section aims to conduct the ablation studies on these two critical modules. To demonstrate the value of the novel \emph{PointMixer} module, we replace the proposed feature extraction network \emph{PointMixer}  with two other recent works, PointNet++ \cite{qi2017pointnet2}, and DGCNN \cite{wang2019dynamic} since both of them can maintain local and global structure information for each point. Two experiments are conducted to demonstrate the better performance of the proposed feature extraction network \textit{PointMixer}. Firstly, the registration performance is evaluated to demonstrate the ability in solving the registration problem. Secondly, the chamfer distance (CD) and mean square error (MSE) between original and generated point clouds are evaluated to demonstrate the point cloud generative quality.

\subsubsection{PointMixer: registration quality}

Table \ref{othernet} shows the comparison of registration performance with other feature learning networks. We find that the proposed \textit{PointMixer} obtains much better registration accuracy than the compared feature learning networks. The reason is that the proposed backbone maintains both global and local structure information for each point at multiple levels. However, the existing methods only use a simple way to concatenate the global structure to each local feature.  
\begin{table}[h]
	\begin{center}
		%\begin{tabular}{p{1.5cm}p{1cm}p{1cm}p{1cm}p{1cm}p{1cm}p{1cm}}
		\begin{tabular}{p{2.4cm}|p{0.8cm}p{0.8cm}p{0.8cm}|p{0.8cm}}\hline		 
			Method      & CD   & RE    & TE &Time   \\ \hline
			% FPFH        &14.113 &0.055  &0.105   &48.50\% &0.02\\ \hline 
			PointNet++ \cite{qi2017pointnet2}  &0.116    &12.84 &0.084 &0.04 \\  
			DGCNN \cite{wang2019dynamic}	    &0.105     &14.11 &0.062  &\bf0.03\\ \hline
			PointMixer  &\bf0.029 &\bf 1.565  &\bf0.006 &0.05  \\ \hline
		\end{tabular}
	\end{center}
	\caption{Ablation study on PointMier's registration quality.}
	\label{othernet}
\end{table}

\subsubsection{PointMixer: generative quality}
Finally, the mean square error (MSE) between the generated $A_g \in R^{N\times 3}$ and transformed original $A_T \in R^{N\times 3}$ point clouds is calculated to further evaluate the point cloud generative quality, $mse = \sum_{i=1}^{N} (A_T^i - A_g^i)^2$, $A_T^i$ and $A_g^i \in R^{1\times 3}$.  The transformation is estimated using the equation \ref{eqform}.  

Table \ref{generation} shows the comparison of generative quality compared with other feature learning networks. We find that the proposed method obtains higher accurate generative accuracy. This experiment demonstrates that the proposed PointMixer works better in the point cloud generation. 

\begin{table}[h]
	\begin{center}
		%\begin{tabular}{p{1.5cm}p{1cm}p{1cm}p{1cm}p{1cm}p{1cm}p{1cm}}
		\begin{tabular}{p{2.4cm}|p{1.2cm}|p{1.2cm}}\hline			 
			Method      &CD  & MSE  \\ \hline
			%FPFH        &0.069 &30.75\% &3.18\\ \hline
			PointNet++ \cite{qi2017pointnet2}  &0.0871 &0.0389 \\ 
			DGCNN \cite{wang2019dynamic} 	    &0.0798 &0.0394 \\ \hline
			PointMixer  &\bf 0.0651 &\bf 0.0181 \\ \hline
		\end{tabular}
	\end{center}
	\caption{Ablation study on PointMixer's generative quality.}
	\label{generation}
\end{table}

\begin{figure}[h]
	\includegraphics[width=\linewidth,height=4.5cm]{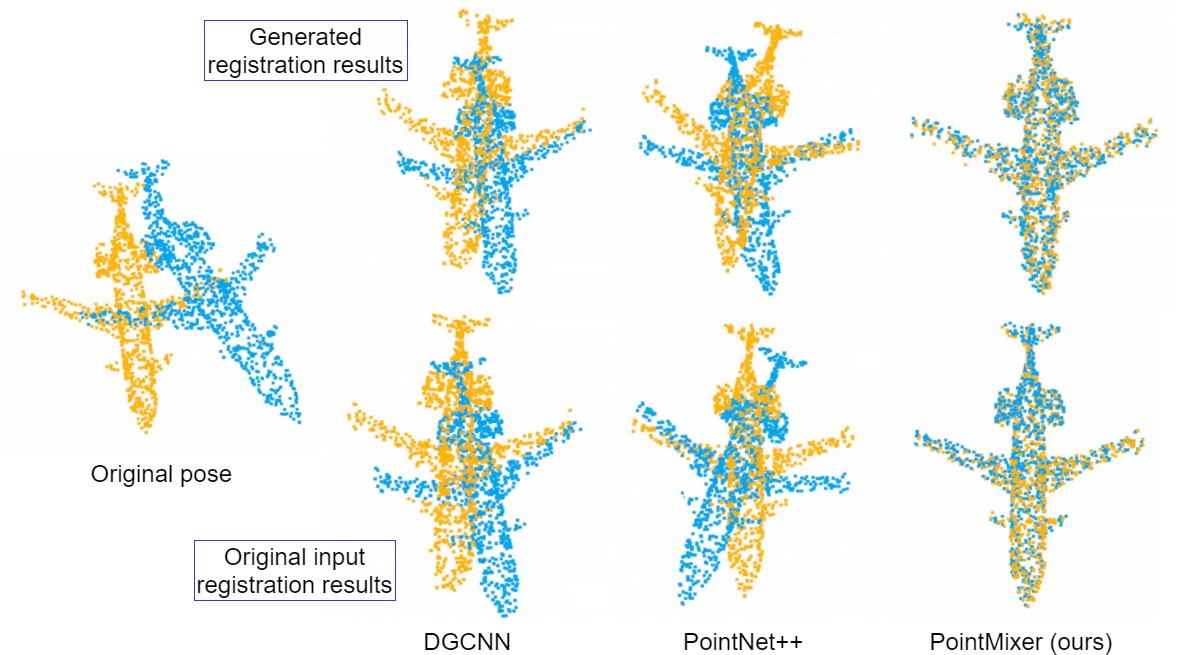}
	\caption{Visual comparison of generated and registration results.}
	\label{generate_fig}
\end{figure}

Figure \ref{generate_fig} shows one example of generated and registration result comparison. They visually show the higher generated and registration quality.

\subsubsection{PDSAC: registration quality}
To demonstrate the value of the proposed PDSAC, we replace the PDSAC with RANSAC \cite{zhou2018open3d} and compare their registration quality. Table \ref{dsac} shows the comparison results. We find that the proposed PDSAC obtains better accuracy than the widely used RANSAC. In addition, the proposed PDSAC achieves more than two times faster than RANSAC.  The reason is that the proposed PDSAC utilizes the matrix operation to conduct the transformation estimation parallel.  However, the RANSAC conducts the sample check iteratively.
\begin{table}[h]
	\begin{center}
		%\begin{tabular}{p{1.5cm}p{1cm}p{1cm}p{1cm}p{1cm}p{1cm}p{1cm}}
		\begin{tabular}{p{2.2cm}|p{0.8cm}p{0.8cm}|p{0.8cm}}\hline		 
			Method       & RE    & TE  &Time  \\ \hline
			% FPFH        &14.113 &0.055  &0.105   &48.50\% &0.02\\ \hline 
			RANSAC \cite{zhou2018open3d}   &1.661  &0.007 &0.25 \\  
			PDSAC    &\bf 1.565  &\bf0.006 &\bf0.11 \\ \hline
		\end{tabular}
	\end{center}
	\caption{Ablation study of PDSAC.}
	\label{dsac}
\end{table}

\subsection{Registration performance on 7Scene}
To evaluate the proposed method on a real-world dataset, following \cite{huang2020feature}, we evaluate the performance on 7scene. Table \ref{7scene} shows that the proposed algorithm also obtains better registration accuracy and faster registration speed than the compared methods. 
\begin{table}[h]
	\begin{center}
		%\begin{tabular}{p{1.5cm}p{1cm}p{1cm}p{1cm}p{1cm}p{1cm}p{1cm}}
		\begin{tabular}{p{2.2cm}|p{0.8cm}p{0.8cm}p{0.8cm}|p{0.8cm}}\hline			 
			Method       & CD   & RE    & TE  &Time\\ \hline
			ICP \cite{besl1992method} 	     &0.118  &7.686  &0.121 &0.41  \\
			FGR \cite{zhou2016fast} 	     &0.051  &1.923 &0.009  &0.21 \\ 
			%DeepGMR      &2.274  &0.015  &0.050 &56.52\% &0.11\\
			RGM \cite{fu2021robust}         &0.044 & 1.797  &0.008 &0.77\\ \hline
			% FMR          &0.038 &96.50\% &0.12 &0.813  &0.005  \\ \hline
			PntLK2\cite{li2021pointnetlk}      &0.058  &2.660  &0.018 &0.11\\ \hline
			GenReg       &\bf0.021  &\bf1.695  &\bf0.008 &\bf0.05  \\ \hline
		\end{tabular}
	\end{center}
	\caption{Registration performance on 7scenes.}
	\label{7scene}
\end{table}

\subsection{Cross-source point cloud registration}
\begin{figure}[h]
	\includegraphics[width=\linewidth,height=2.5cm]{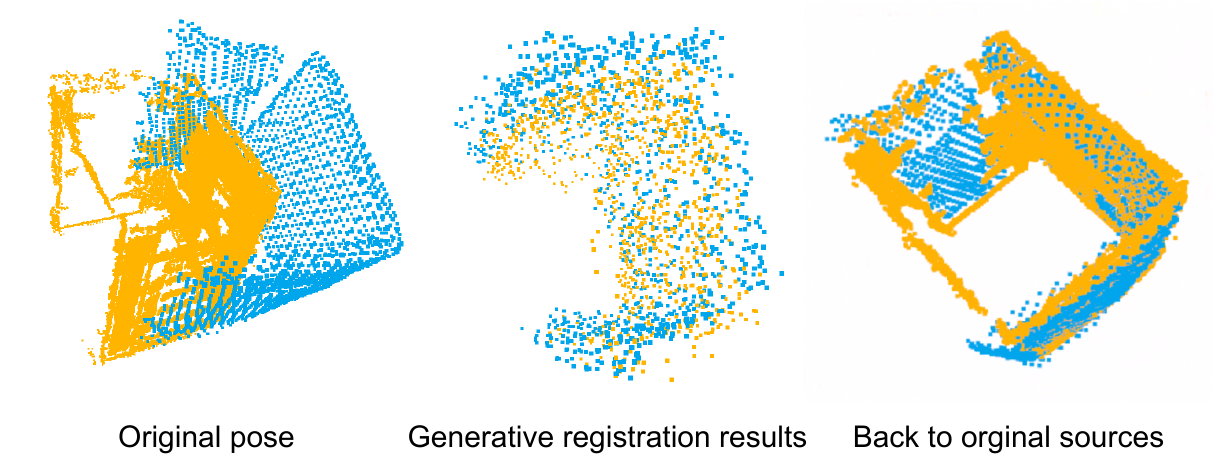}
	\caption{Registration result of cross-source point clouds.}
	\label{cs_fig}
\end{figure}

Recently, point cloud registration from heterogeneous sensors has attracted more research attention. We use our trained model from ModelNet40 to test on the cross-source point clouds from Kinect and RGB camera (reconstructed using VSFM \cite{wu2011visualsfm}) \cite{huang2021comprehensive}. Figure \ref{cs_fig} shows one example that the proposed algorithm can achieve accurate registration results on the cross-source point clouds. 

\section{Conclusion}
We propose the first generative-based registration method to generate the registration results directly. This simple and effective method demonstrates high registration accuracy and much faster speed in the experiments. Our method contains the new PointMixer and FI modules to efficiently maintain the global and local structure information at multiple levels from self and cross point clouds. We also design a parallel and differential sample consensus method to estimate the transformation matrix of original point clouds based on the generated aligned results in an end-to-end manner. Several loss functions are proposed to train our method to learn the implicit data distribution from inputs and generate the aligned point clouds. The possible   future work is to extend this method to solve other tasks like optical flow and localization.

%%%%%%%%% REFERENCES
{\small
\bibliographystyle{ieee_fullname}
\bibliography{egbib}
}

\clearpage
\setcounter{section}{0}
\setcounter{figure}{0}

\section{Appendix}
In this supplementary material, we first provide detailed network architectures and training details. Then, broader impact is discussed.

\subsection{Network Architecture}
Firstly, the T-Net contains three-layers of MLP and a feature interaction module. The three-layer MLP aims to extract local feature for each point cloud. The channels are [64,128,1024]. Secondly, a max-pooling is applied to get the global feature. Thirdly, the interaction module concatenates the both global features, and uses the concatenated feature into a three-layer MLP to estimate the six transformation parameters. The channels of the three-layer MLP in the interaction module are [512,256,6].

\begin{figure}[h]
	\includegraphics[width=\linewidth]{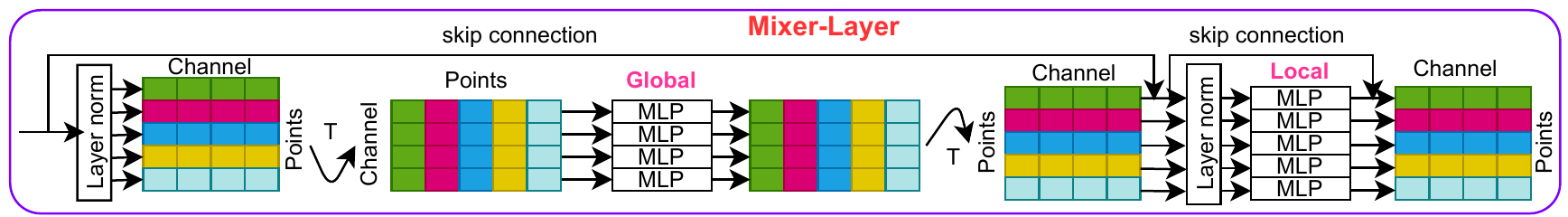}
	\caption{Network details of Mixer layer. }
	\label{mixer}
\end{figure}

Secondly, the proposed PointMixer contains two Mixer layers. The details of mixer layer is shown in Figure \ref{mixer}. Each mixer layers conduct Token-mixing MLP on point dimension first and followed by channel-mixing MLP on channel dimension. Each MLP contains two fully connected layers. The neurons are [1024,1024] and [128,128] for the first and second mixer layers, respectively. The token-mixing obtains global feature. This global feature then passes to channel-mixing for local feature extraction. By this mechanism, the proposed mixer layer can maintain both global and local information for each point.

\begin{figure}[h]
	\includegraphics[width=\linewidth]{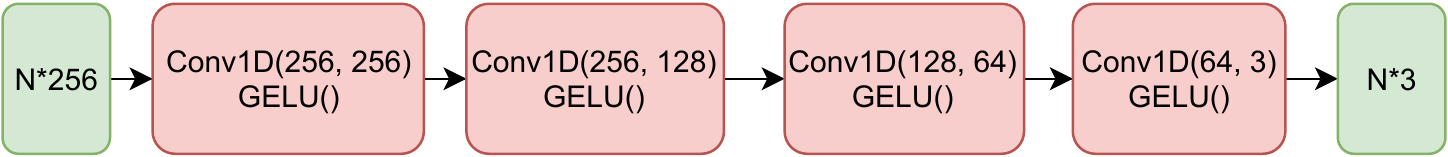}
	\caption{The details of decoder neural network.}
	\label{decoder}
\end{figure}

Thirdly, the details of decoder neural network are shown in Figure \ref{decoder}. It contains four 1D convolution layer and activates by GELU.

\begin{figure}[h]
	\includegraphics[width=\linewidth]{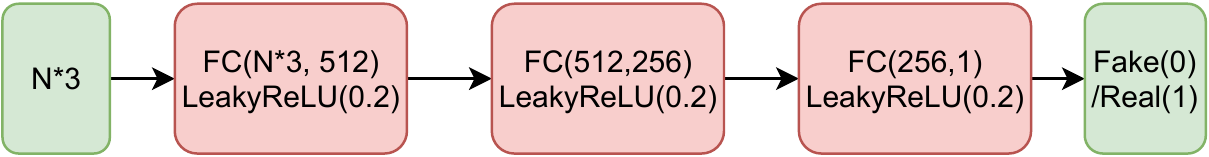}
	\caption{The details of discriminator neural network.}
	\label{discriminator}
\end{figure}

Finally, the adversarial network consists of three fully connected layers and each layer is activated by LeakyReLU. Figure \ref{discriminator} shows the details.

\subsection{Training details}
The proposed method is trained and evaluated on Quadro RTX 6000 (24GB). The neural network is trained 200 epochs, which takes about 20 hours in total. The batch size is set to 16 which occupied about 16GB GPU memory. During the network training of GAN, the discriminator is not easy to reach convergent. We train the generative network every 5 steps while train the discriminator every step.

\subsection{More visual results }
\begin{figure}[h]
	\includegraphics[width=\linewidth]{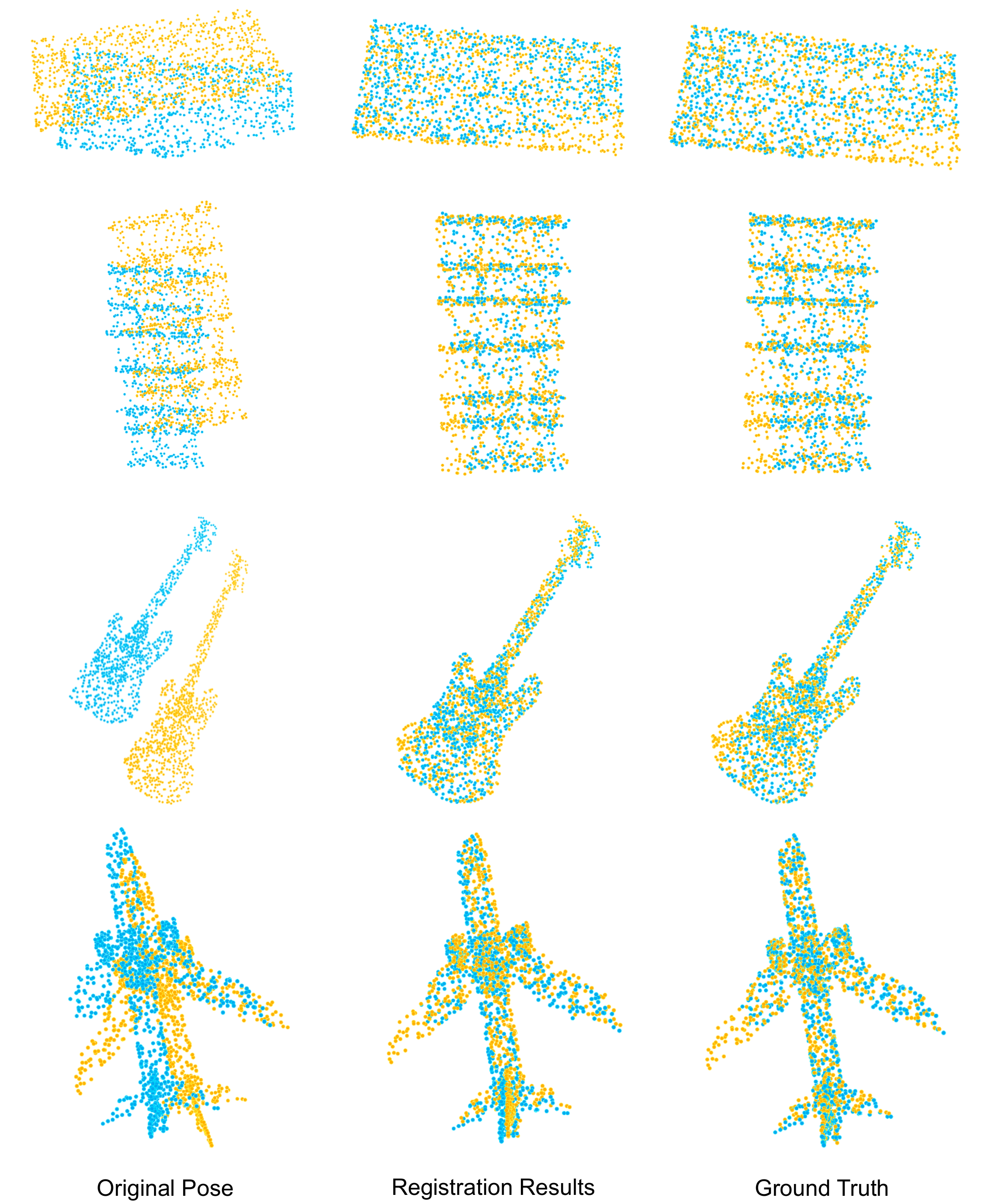}
	\caption{More visual results on partial overlap problem.}
	\label{partial_sup}
\end{figure}

Figure \ref{partial_sup} shows more visual results on partial overlap problem.

\subsection{Discussion}
\textbf{Broader impact} We present a new deep generative method for point cloud registration. It makes a first attempt towards the aligned point clouds generation. This work opens a new research direction to leverage the available big data to directly learn how to generate the aligned point clouds by neural networks. Our work can contribute to a wide range of applications to achieve both high efficiency and accuracy, such as 3D matching, 3D search, augment reality and simultaneous localization and mapping (SLAM), or any other where point cloud registration plays a role. As our method aims at directly generating the registration results, it achieves high efficiency and accuracy. Our method meets the real-time registration requirement. 

\end{document}